\documentclass[11pt,a4paper]{article}
\usepackage[hyperref]{acl2019}
\usepackage{times}
\usepackage{latexsym}
\usepackage{amssymb}
\usepackage{amsmath}

\usepackage[pdftex]{graphicx}

\usepackage{stfloats}

\usepackage{url}

\aclfinalcopy 
\def\aclpaperid{89} 

\title{The CUED's Grammatical Error Correction Systems for BEA-2019}

\author{Felix Stahlberg \and Bill Byrne\\
 Department of Engineering\\
 University of Cambridge\\
 Trumpington St, Cambridge CB2 1PZ, UK  \\
  {\tt \{fs439,wjb31\}@cam.ac.uk} \\
  }

\date{}

\begin{document}
\maketitle
\begin{abstract}
  We describe two entries from the Cambridge University Engineering Department to the BEA 2019 Shared Task on grammatical error correction. Our submission to the low-resource track is based on prior work on using finite state transducers together with strong neural language models. Our system for the restricted track is a purely neural system consisting of neural language models and neural machine translation models trained with back-translation and a combination of checkpoint averaging and fine-tuning -- without the help of any additional tools like spell checkers. The latter system has been used inside a separate system combination entry in cooperation with the Cambridge University Computer Lab.
\end{abstract}

\section{Introduction}

The automatic correction of errors in text [\textit{In a such situaction} $\rightarrow$ \textit{In such a situation}] is receiving more and more attention from the natural language processing community. A series of competitions has been devoted to grammatical error correction (GEC): the CoNLL-2013 shared task~\citep{conll2013}, the CoNLL-2014 shared task~\citep{conll2014}, and finally the BEA 2019 shared task~\citep{bea19}. This paper presents the contributions from the Cambridge University Engineering Department to the latest GEC competition at the BEA 2019 workshop.

We submitted systems to two different tracks. The {\em low-resource track} did not permit the use of parallel training data except a small development set with around 4K sentence pairs. For our low-resource system we extended our prior work on finite state transducer based GEC~\citep{fst-gec} to handle new error types such as punctuation errors as well as insertions and deletions of a small number of frequent words. For the {\em restricted track}, the organizers provided 1.2M pairs (560K without identity mappings) of corrected and uncorrected sentences. Our goal on the restricted track was to explore the potential of purely neural models for grammatical error correction.\footnote{Models will be published at \url{http://ucam-smt.github.io/sgnmt/html/bea19_gec.html}.}  We confirm the results of \citet{backtranslation-gec} and report substantial gains by applying back-translation~\citep{backtranslation} to GEC -- a data augmentation technique common in machine translation. Furthermore, we noticed that large parts of the training data do not match the target domain. We mitigated the domain gap by over-sampling the in-domain training corpus, and by fine-tuning through continued training. Our final model is an ensemble of four neural machine translation (NMT) models and two neural language models (LMs) with Transformer architecture~\citep{transformer}. Our purely neural system was also part of the joint submission with the Cambridge University Computer Lab described by \citet{bea19-cled}.

\section{Low-resource Track Submission}

\subsection{FST-based Grammatical Error Correction}

\begin{figure*}[t!]
\centering
\small
\includegraphics[width=0.82\linewidth]{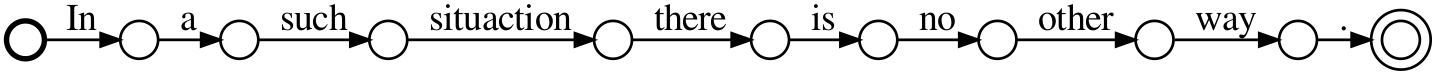}
\caption{Input FST $I$ representing the source sentence `In a such situaction there is no other way.'. We follow standard convention and highlight the start state in bold and the final state with a double circle.}
\label{fig:input-fst}
\end{figure*}

\citet{fst-gec} investigated the use of finite state transducers (FSTs) for neural grammatical error correction. They proposed a cascade of FST compositions to construct a hypothesis space which is then rescored with a neural language model. We will outline this approach and explain our modifications in this section. For more details we refer to \citep{fst-gec}.

In a first step, the source sentence is converted to an FST $I$ (Fig.~\ref{fig:input-fst}). This initial FST is augmented by composition (denoted with the $\circ$-operator) with various other FSTs to cover different error types. Composition is a widely used standard operation on FSTs and supported efficiently by FST toolkits such as OpenFST~\citep{openfst}. We construct the hypothesis space as follows:\footnote{Note that our description differs from \citep{fst-gec} in the following ways: First, we use additional FSTs to allow insertions and deletions. Second, we integrate penalties directly into the FSTs rather than using special tokens in combination with a penalization transducer.}

\begin{figure}[t!]
\centering
\small
\includegraphics[width=0.4\linewidth]{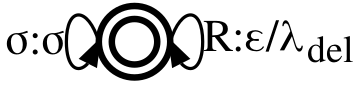}
\caption{Deletion FST $D$ which can map any token in the list $R$ from Tab.~\ref{tab:deletion-list} to $\epsilon$. The $\sigma$-label matches any symbol and maps it to itself.}
\label{fig:deletion-fst}
\end{figure}

\begin{table}
\centering
\small
\begin{tabular}{|c|c|}
\hline
\textbf{Deletion Frequency} & \textbf{Token} \\
\textbf{(dev set)} & \\
\hline
164 & the \\
78 & , \\
50 & a \\
33 & to \\
20 & it \\
18 & of \\
16 & in \\
12 & that \\
8 & will \\
8 & have \\
8 & for \\
8 & an \\
7 & is \\
7 & - \\
6& they \\
6 & 's \\
6 & and \\
5 & had \\
\hline
\end{tabular}
\caption{List of tokens $R$ that can be deleted by the deletion transducer $D$ in Fig.~\ref{fig:deletion-fst}.}\label{tab:deletion-list}
\end{table}

\begin{figure}[t!]
\centering
\small
\includegraphics[width=0.7\linewidth]{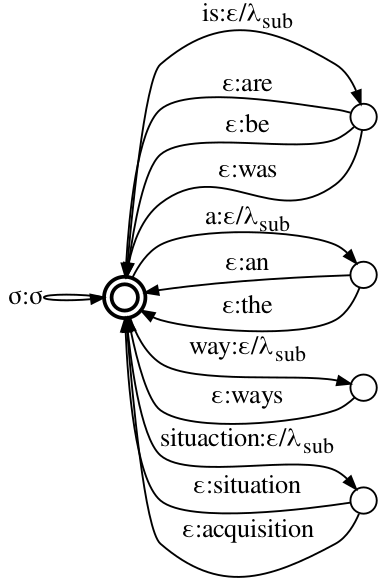}
\caption{Edit FST $E$ which allows substitutions with a cost of $\lambda_\text{sub}$. The $\sigma$-label matches any symbol and maps it to itself at no cost.}
\label{fig:edit-fst}
\end{figure}

\begin{figure}[t!]
\centering
\small
\includegraphics[width=0.68\linewidth]{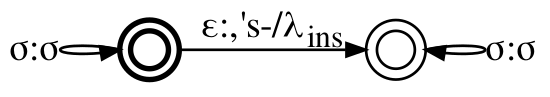}
\caption{Insertion FST $A$ for adding the symbols ``,'', ``-'', and ``'s'' at a cost of $\lambda_\text{ins}$. The $\sigma$-label matches any symbol and maps it to itself at no cost.}
\label{fig:add-fst}
\end{figure}

\begin{table*}
\centering
\small
\begin{tabular}{|ccc|cc|ccc|ccc|}\hline
{\bf Sub} & {\bf Del} & {\bf Ins} & {\bf LM} & {\bf Beam} & \multicolumn{3}{c|}{{\bf CoNLL-2014}} & \multicolumn{3}{c|}{{\bf BEA-2019 Dev}} \\ 
 &  &  &  & & {\bf P} & {\bf R} & {\bf M2} & {\bf P} & {\bf R} & {\bf ERRANT} \\ \hline
\multicolumn{5}{|l|}{Best published: \citet{fst-gec}} & 54.12 & 25.52 & 44.21 & \multicolumn{3}{c|}{n/a} \\ \hline
\checkmark & & & 1x & 8 & 58.59 & 24.14  & 45.58 & 42.44 &	14.68 &	30.79 \\
\checkmark & \checkmark & & 1x & 8 &  59.01  & 26.07 & 47.11 & 41.21 &	16.47 &	31.69 \\
\checkmark & \checkmark & \checkmark & 1x & 8 & 52.89 & 26.68  & 44.20 & 40.09	& 19.97	& 33.36 \\
\checkmark & \checkmark & \checkmark & 2x & 8 & 54.05 & 26.71  & 44.87 & 40.70   & 20.01  & 33.73 \\
\checkmark & \checkmark & \checkmark & 2x & 16 & 57.05 & 27.22  & 46.80 & 42.02 & 19.76 & 34.29 \\
\checkmark & \checkmark & \checkmark & 2x & 32 & 58.48 & 28.21  & 48.15 & 42.37	& 19.92	& 34.58 \\
\hline
\end{tabular}
\caption{Results on the low-resource track. The $\lambda$-parameters are tuned on the BEA-2019 dev set.}\label{tab:results-low-resource}
\end{table*}

\begin{enumerate}
    \item We compose the input $I$ with the deletion transducer $D$ in Fig.~\ref{fig:deletion-fst}. $D$ allows to delete tokens on the short list shown in Tab.~\ref{tab:deletion-list} at a cost $\lambda_\text{del}$. We selected $R$ by looking up all tokens which have been deleted in the dev set more than five times and manually filtered that list slightly. We did not use the full list of dev set deletions to avoid under-estimating $\lambda_\text{del}$ in tuning.
    \item In a next step, we compose the transducer from step 1 with the edit transducer $E$ in Fig.~\ref{fig:edit-fst}. This step addresses substitution errors such as spelling or morphology errors. Like \citet{fst-gec}, we use the confusion sets of \citet{chris-lm} based on CyHunspell for spell checking \citep{cyhunspell}, the AGID morphology database for morphology errors \citep{agid}, and manually defined corrections for determiner and preposition errors to construct $E$. Additionally, we extracted all substitution errors from the BEA-2019 dev set which occurred more than five times, and added a small number of manually defined rules that fix tokenization around punctuation symbols.
    \item We found it challenging to allow insertions in LM-based GEC because the LM often prefers inserting words with high unigram probability such as articles and prepositions before less predictable words like proper names. We therefore restrict insertions to the three tokens ``,'', ``-'', and ``'s'' and allow only one insertion per sentence. We achieve this by adding the transducer $A$ in Fig.~\ref{fig:add-fst} to our composition cascade.
    \item Finally, we map the word-level FSTs to the subword-level by composition with a mapping transducer $T$ that applies byte pair encoding \citep[BPE]{bpe} to the full words. Word-to-BPE mapping transducers have been used in prior work to combine word-level models with subword-level neural sequence models~\citep{fst-gec,sgnmt1,sgnmt2,mbr-nmt}.
\end{enumerate}

In a more condensed form, we can describe the final transducer as:
\begin{equation}
\label{eq:fst-cascade}
I\circ D \circ E \circ A \circ T
\end{equation}
with $D$ for deletions, $E$ for substitutions, $A$ for insertions, and $T$ for converting words to BPE tokens. Path scores in the FST in Eq.~\ref{eq:fst-cascade} are the accumulated penalties $\lambda_\text{del}$, $\lambda_\text{sub}$, and $\lambda_\text{ins}$. The $\lambda$-parameters are tuned on the dev set using a variant of Powell search~\citep{powell}. We apply standard FST operations like output projection, $\epsilon$-removal, determinization, minimization, and weight pushing~\citep{mohri-lang,mohri-push} to help downstream decoding. Following \citet{fst-gec} we then use the resulting transducer to constrain a neural LM beam decoder.

\subsection{Experimental Setup}
\label{sec:low-resource-setup}

Our LMs are Transformer~\citep{transformer} decoders (\texttt{transformer\_big}) trained using the Tensor2Tensor library~\citep{t2t}. We delay SGD updates~\citep{ucam-wmt18,danielle-syntax} with factor 2 to simulate 500K training steps with 8 GPUs on 4 physical GPUs. Training batches contain about 4K source and target tokens. Our LM training set comprises the monolingual {\em news2015}-{\em news2018} English training sets\footnote{\url{http://www.statmt.org/wmt19/translation-task.html}} from the WMT evaluation campaigns~\citep{wmt18} after language detection~\citep{langdetect} (138M sentences) and subword segmentation using byte pair encoding~\citep{bpe} with 32K merge operations. For decoding, we use our SGNMT tool~\citep{sgnmt1,sgnmt2} with OpenFST backend~\citep{openfst}.

\subsection{Results}

We report M2~\citep{m2} scores on the CoNLL-2014 test set~\citep{conll2014} and span-based ERRANT scores~\citep{errant} on the BEA-2019 dev set~\citep{bea19}. On CoNLL-2014 we compare with the best published results with comparable amount of parallel training data. We refer to \citep{bea19} for a full comparison of BEA-2019 systems. We tune our systems on BEA-2019 and only report the performance on CoNLL-2014 for comparison to prior work.

Tab.~\ref{tab:results-low-resource} summarizes our low-resource experiments. Our substitution-only system already outperforms the prior work of \citet{fst-gec}. Allowing for deletions and insertions improves the ERRANT score on BEA-2019 Dev by 2.57 points. We report further gains on both test sets by ensembling two language models and increasing the beam size.

\subsection{Differences Between CoNLL-2014 and BEA-2019 Dev}
\label{sec:diff-conll-bea}

Our results in Tab.~\ref{tab:results-low-resource} differ significantly between the CoNLL-2014 test set and the BEA-2019 dev set. Allowing insertions is beneficial on BEA-2019 Dev but decreases the M2 score on CoNLL-2014. Increasing the beam size improves our system by 3.28 points on CoNLL-2014 while the impact on BEA-2019 Dev is smaller (+0.85 points). These differences can be partially explained by comparing the error type frequencies in the reference annotations in both test sets (Tab.~\ref{tab:conll-vs-bea}). Samples in CoNLL-2014 generally need more corrections per sentence than in BEA-2019 Dev. More importantly, the CoNLL-2014 test set contains fewer missing words, but much more unnecessary words than BEA-2019 Dev. This mismatch tempers with tuning as we explicitly tune insertion and deletion penalties.

\begin{table}
\centering
\small
\begin{tabular}{|l|r|r|r|r|}
\hline
& \multicolumn{2}{c|}{{\bf Per Sentence}}  & \multicolumn{2}{c|}{{\bf Per Word}} \\
& {\bf CoNLL} & {\bf BEA} & {\bf CoNLL} & {\bf BEA} \\ \hline
Missing & 0.35 & 0.46 & 1.51\% & 2.30\% \\
Replacement & 1.52 & 1.31 & 6.62\% & 6.57\% \\
Unnecessary & 0.42 & 0.19 & 1.83\% & 0.98\% \\ \hline
Total & 2.29 & 1.96 & 9.95\% & 9.86\% \\
\hline
\end{tabular}
\caption{Number of correction types in CoNLL-2014 and BEA-2019 Dev references.}\label{tab:conll-vs-bea}
\end{table}









\section{Restricted Track Submission}

In contrast to our low-resource submission, our restricted system entirely relies on neural models and does not use any external NLP tools, spell checkers, or hand-crafted confusion sets. For simplicity, we also chose to use standard implementations~\citep{t2t} of standard Transformer~\citep{transformer} models with standard hyper-parameters. This makes our final system easy to deploy as it is a simple ensemble of standard neural models with minimal preprocessing (subword segmentation). Our contributions on this track focus on NMT training techniques such as over-sampling, back-translation, and fine-tuning. We show that over-sampling effectively reduces domain mismatch. We found back-translation~\citep{backtranslation} to be a very effective technique to utilize unannotated training data. However, while over-sampling is commonly used in machine translation to balance the number of real and back-translated training sentences, we report that using over-sampling this way for GEC hurts performance. Finally, we propose a combination of checkpoint averaging~\citep{ckpt-avg} and continued training to adapt our NMT models to the target domain.

\subsection{Experimental Setup}

\begin{table}
\centering
\small
\begin{tabular}{|l|l|l|}
\hline
& {\sc Base} & {\sc Big} \\ \hline
T2T HParams set & \texttt{trans.\_base} & \texttt{trans.\_big} \\
\# physical GPUs & 4 & 4 \\
Batch size & 4,192 & 2,048 \\
SGD delay factor & 2 & 4 \\
\# training iterations & 300K & 400K \\
Beam size & 4 & 8 \\
\hline
\end{tabular}
\caption{NMT setups {\sc Base} and {\sc Big} used in our experiments for the restricted track.}\label{tab:trans-setups}
\end{table}

\begin{table}
\centering
\small
\begin{tabular}{|l|r|r|}
\hline
& \multicolumn{2}{c|}{{\bf Number of Sentences}} \\
& {\bf With Identities} & {\bf W/o Identities} \\ \hline
FCE & 28K & 18K \\
Lang-8 & 1,038K & 498K \\
NUCLE & 57K & 21K \\
W\&I+LOCNESS & 34K & 23K \\ \hline
{\bf Total} & {\bf 1,157K} & {\bf 560K} \\
\hline
\end{tabular}
\caption{BEA-2019 parallel training data with and without removing pairs where source and target sentences are the same.}\label{tab:train-corpus}
\end{table}

We use neural LMs and neural machine translation (NMT) models in our restricted track entry. Our neural LM is as described in Sec.~\ref{sec:low-resource-setup}. Our LMs and NMT models share the same subword segmentation. We perform exploratory NMT experiments with the {\sc Base} setup, but switch to the {\sc Big} setup for our final models. Tab.~\ref{tab:trans-setups} shows the differences between both setups. Tab.~\ref{tab:train-corpus} lists some corpus statistics for the BEA-2019 training sets. In our experiments without fine-tuning we decode with the average of the 20 most recent checkpoints~\citep{ckpt-avg}. We use the SGNMT decoder~\citep{sgnmt1,sgnmt2} in all our experiments.

\paragraph{In-domain corpus over-sampling}

\begin{table*}
\centering
\small
\begin{tabular}{|c|c|ccc|ccc|}\hline
{\bf W\&I+LOCNESS} & {\bf Ratio} & \multicolumn{3}{c|}{{\bf CoNLL-2014}} & \multicolumn{3}{c|}{{\bf BEA-2019 Dev}} \\ 
{\bf Over-sampling Rate} &  & {\bf P} & {\bf R} & {\bf M2} & {\bf P} & {\bf R} & {\bf ERRANT} \\ \hline
 1x & 1:33 & 59.88 & 17.46  & 40.30 & 38.20 &   15.09 &  29.24 \\
 4x & 1:8 & 59.16 & 17.20  & 39.76 & 40.40 & 16.67 & 31.44 \\
 8x & 1:4 & 57.73 & 17.76  & 39.81 & 39.19 & 16.73 & 30.90 \\
\hline
\end{tabular}
\caption{Over-sampling the BEA-2019 in-domain corpus W\&I+LOCNESS under {\sc Base} models. The second column contains the ratio of W\&I+LOCNESS samples to training samples from the other corpora.}\label{tab:results-locness-os}
\end{table*}

The BEA-2019 training corpora (Tab.~\ref{tab:train-corpus}) differ significantly not only in size but also their closeness to the target domain. The W\&I+LOCNESS corpus is most similar to the BEA-2019 dev and test sets in terms of domains and the distribution over English language proficiency, but only consists of 34K sentence pairs. To increase the importance of in-domain training samples we over-sampled the W\&I+LOCNESS corpus with different rates. Tab.~\ref{tab:results-locness-os} shows that over-sampling by factor 4 (i.e.\ adding the W\&I+LOCNESS corpus four times to the training set) improves the ERRAMT $F_{0.5}$-score by 2.2 points on the BEA-2019 dev set and does not lead to substantial losses on the CoNLL-2014 test set. We will over-sample the W\&I+LOCNESS corpus by four in all subsequent experiments.

\paragraph{Removing identity mappings}

\begin{table}
\centering
\small
\begin{tabular}{|@{\hspace{0.2em}}c@{\hspace{0.2em}}|@{\hspace{0.4em}}c@{\hspace{1em}}c@{\hspace{1em}}c@{\hspace{0.4em}}|@{\hspace{0.4em}}c@{\hspace{1em}}c@{\hspace{1em}}c@{\hspace{0.4em}}|}\hline
{\bf Identity} & \multicolumn{3}{@{\hspace{0em}}c@{\hspace{0em}}}{{\bf CoNLL-2014}} & \multicolumn{3}{c|}{{\bf BEA-2019 Dev}} \\ 
{\bf Removal}  & {\bf P} & {\bf R} & {\bf M2} & {\bf P} & {\bf R} & {\bf ERR.} \\ \hline
$\times$  & 59.16 & 17.20  & 39.76 & 40.40 & 16.67 & 31.44 \\
 \checkmark & 53.34 & 28.83  & 45.59 & 33.04  & 23.14 & 30.44 \\
\hline
\end{tabular}
\caption{Impact of identity removal on {\sc Base} models.}\label{tab:results-id}
\end{table}

Previous works often suggested to remove unchanged sentences (i.e.\ source and target sentences are equal) from the training corpora~\citep{fst-gec,gec-pretraining,marcin2018}. We note that removing these identity mappings can be seen as measure to control the balance between precision and recall. As shown in Tab.~\ref{tab:results-id}, removing identities encourages the model to make more corrections and thus leads to higher recall but lower precision. It depends on the test set whether this results in an improvement in $F_{0.5}$ score. For the subsequent experiments we found that removing identities in the parallel training corpora but not in the back-translated synthetic data works well in practice.

\paragraph{Back-translation}

\begin{table*}[!b]
\centering
\small
\begin{tabular}{|c|c|c|ccc|ccc|}\hline
{\bf Over-sampling Rate} & {\bf Number of} & {\bf Ratio} & \multicolumn{3}{c|}{{\bf CoNLL-2014}} & \multicolumn{3}{c|}{{\bf BEA-2019 Dev}} \\ 
{\bf (Real Data)} & {\bf Synthetic Sentences} &  & {\bf P} & {\bf R} & {\bf M2} & {\bf P} & {\bf R} & {\bf ERRANT} \\ \hline
 1x & 0 & - & 53.34 & 28.83  & 45.59 & 33.04  & 23.14 & 30.44 \\
 1x & 1M & 1:1.6 & 56.17 & 31.30  & 48.47 & 37.79 &	23.86 &	33.84 \\
 1x & 3M & 1:4.8 & 61.40 & 34.29  & 53.02 & 42.62 &	25.30 &	37.49 \\
 1x & 5M & 1:7.9 & 64.18 & 34.27  & 54.64 & 44.69 &	25.59 &	38.88 \\ \hline
 3x & 3M & 1:1.6 & 57.12 & 32.55  & 49.63 & 40.08 &	24.79 &	35.68 \\
 6x & 5M & 1:1.3 & 59.15 & 33.99  & 51.52 & 41.52 &	25.05 &	36.69 \\
\hline
\end{tabular}
\caption{Using back-translation for GEC ({\sc Base} models). The third column contains the ratio between real and synthetic sentence pairs.}\label{tab:results-backtrans}
\end{table*}

Back-translation~\citep{backtranslation} has become the most widely used technique to use monolingual data in neural machine translation. Back-translation extends the existing parallel training set by additional training samples with real English target sentences but synthetic source sentences. Different methods have been proposed to synthesize the source sentence such as using dummy tokens~\citep{backtranslation}, copying the target sentence~\citep{copytarget}, or sampling from or decoding with a reverse sequence-to-sequence model~\citep{backtranslation,backtranslation-fb,backtranslation-gec}. The most popular approach is to generate the synthetic source sentences with a reverse model that is trained to transform target to source sentences using beam search. In GEC, this means that the reverse model learns to introduce errors into a correct English sentence. Back-translation has been applied successfully to GEC by~\citet{backtranslation-gec}. We confirm the effectiveness of back-translation in GEC and discuss some of the differences between applying this technique to grammatical error correction and machine translation.

Our experiments with back-translation are summarized in Tab.~\ref{tab:results-backtrans}. Adding 1M synthetic sentences to the training data already yields very substantial gains on both test sets. We achieve our best results with 5M synthetic sentences (+8.44 on BEA-2019 Dev). In machine translation, it is important to maintain a balance between authentic and synthetic data~\citep{backtranslation,backtranslation-ana,sys-uedin-wmt16}. Over-sampling the real data is a common practice to rectify that ratio if large amounts of synthetic data are available. Interestingly, over-sampling real data in GEC hurts performance (row 3 vs.\ 5 in Tab.~\ref{tab:results-backtrans}), and it is possible to mix real and synthetic sentences at a ratio of 1:7.9 (last three rows in Tab.~\ref{tab:results-backtrans}). We will proceed with the 5M setup for the remainder of this paper.

\paragraph{Fine-tuning}

\begin{figure}[t!]
\centering
\small
\includegraphics[width=0.93\linewidth]{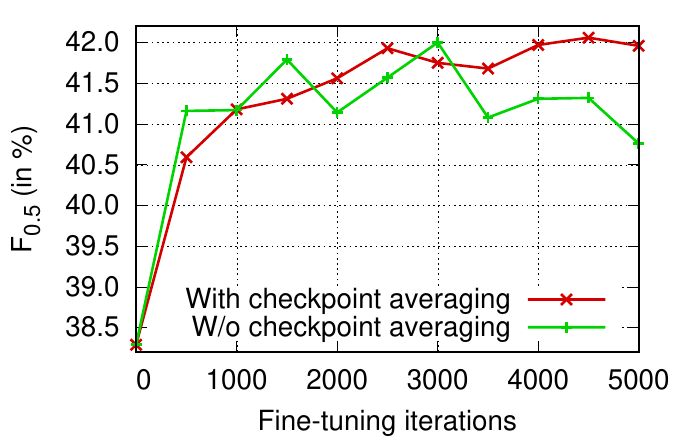}
\caption{Span-based ERRANT $F_{0.5}$ scores on the BEA-2019 dev set over the number of fine-tuning training iterations (single GPU, SGD delay factor~\citep{danielle-syntax} of 16).}
\label{fig:fine-tuning}
\end{figure}

\begin{table*}
\centering
\small
\begin{tabular}{|c|c|ccc|ccc|}\hline
{\bf Fine-tuning} & {\bf Checkpoint} & \multicolumn{3}{c|}{{\bf CoNLL-2014}} & \multicolumn{3}{c|}{{\bf BEA-2019 Dev}} \\ 
{\bf (Continued Training)} & {\bf Averaging} & {\bf P} & {\bf R} & {\bf M2} & {\bf P} & {\bf R} & {\bf ERRANT} \\ \hline
  &  &  63.61 & 33.39  & 53.86 & 44.16 & 25.01 & 38.29 \\
  & \checkmark &  64.18 & 34.27  & 54.64 & 44.69 &	25.59 &	38.88 \\
  \checkmark & &  64.98 & 33.05  & 54.46 & 48.62 & 27.19 & 42.00 \\
  \checkmark & \checkmark &  66.03 & 34.17  & 55.65 & 	48.99 & 26.87 &  42.06 \\
\hline
\end{tabular}
\caption{Fine-tuning through continued training on W\&I+LOCNESS and checkpoint averaging with a {\sc Base} model with 5M back-translated sentences.}\label{tab:results-ft}
\end{table*}

As explained previously, we over-sample the W\&I+LOCNESS corpus by factor 4 to mitigate the domain gap between the training set and the BEA-2019 dev and test sets. To further adapt our system to the target domain, we fine-tune the NMT models on W\&I+LOCNESS after convergence on the full training set. We do that by continuing training on W\&I+LOCNESS from the last checkpoint of the first training pass. Fig.~\ref{fig:fine-tuning} plots the $F_{0.5}$ score on the BEA-2019 dev set for two different setups. For the red curve, we average all checkpoints~\citep{ckpt-avg} (including the last unadapted checkpoint) up to a certain training iteration. Checkpoints are dumped every 500 steps. The green curve does not use any checkpoint averaging. Checkpoint averaging helps to smooth out fluctuations in $F_{0.5}$ score, and also generalizes better to CoNLL-2014 (Tab.~\ref{tab:results-ft}).

\paragraph{Final system}

\begin{table*}[!b]
\centering
\small
\begin{tabular}{|ccc|ccc|ccc|}\hline
{\bf NMT} & {\bf Fine-tuning} & {\bf LM} &  \multicolumn{3}{c|}{{\bf CoNLL-2014}} & \multicolumn{3}{c|}{{\bf BEA-2019 Dev}} \\ 
  &  & & {\bf P} & {\bf R} & {\bf M2} & {\bf P} & {\bf R} & {\bf ERRANT} \\ \hline
\multicolumn{3}{|l|}{Best published: \citet{gec-pretraining}} & 71.57 & 38.65 & 61.15 & \multicolumn{3}{c|}{n/a} \\ \hline
1x & & & 64.04 & 35.74  & 55.28 & 45.86 &	26.46 &	40.00 \\
1x & \checkmark & & 66.57 & 35.21  & 56.50 & 51.57 &	27.49 &	43.88 \\
1x & \checkmark  & 2x & 61.53 & 40.44  & 55.72 & 48.30 & 33.08 & 44.23 \\ \hline
4x & \checkmark  &  & 70.37 & 35.12  & 58.60 & 55.84 &	27.80 &	46.47 \\
4x & \checkmark  & 2x & 66.89 & 39.85  & 58.90 & 53.17 &	32.89 &	47.34 \\
\hline
\end{tabular}
\caption{Final results on the restricted track with {\sc Big} models and back-translation.}\label{tab:results-restricted}
\end{table*}

Tab.~\ref{tab:results-restricted} contains our experiments with the {\sc Big} configuration. In addition to W\&I+LOCNESS over-sampling, back-translation with 5M sentences, and fine-tuning with checkpoint averaging, we report further gains by adding the language models from our low-resource system (Sec.~\ref{sec:low-resource-setup}) and ensembling. Our best system (4 NMT models, 2 language models) achieves 58.9 M2 on CoNLL-2014, which is slightly (2.25 points) worse than the best published result on that test set~\citep{gec-pretraining}. However, we note that we have tailored our system towards the BEA-2019 dev set and not the CoNLL-2013 or CoNLL-2014 test sets. As we argued in Sec.~\ref{sec:diff-conll-bea}, our results throughout this work suggest strongly that the optimal system parameters for these test sets are very different from each other, and that our final system settings are not optimal for CoNLL-2014. We also note that unlike the system of \citet{gec-pretraining}, our system for the restricted track does not use spell checkers or other NLP tools but relies solely on neural sequence models.

\section{Conclusion}

We participated in the BEA 2019 Shared Task on grammatical error correction with submissions to the low-resource and the restricted track. Our low-resource system is an extension of prior work on FST-based GEC~\citep{fst-gec} to allow insertions and deletions. Our restricted track submission is a purely neural system based on standard NMT and LM architectures. We pointed out the similarity between GEC and machine translation, and demonstrated that several techniques which originate from MT research such as over-sampling, back-translation, and fine-tuning, are also useful for GEC. Our models have been used in a joint submission with the Cambridge University Computer Lab~\citep{bea19-cled}.

\section*{Acknowledgments}

This work was supported by the U.K. Engineering and Physical Sciences Research Council (EPSRC) grant EP/L027623/1 and has been performed using resources provided by the Cambridge Tier-2 system operated by the University of Cambridge Research Computing Service\footnote{\url{http://www.hpc.cam.ac.uk}} funded by EPSRC Tier-2 capital grant EP/P020259/1.

\bibliography{acl2019}
\bibliographystyle{acl_natbib}


\end{document}